\documentclass{article}
\usepackage{spconf,amsmath,graphicx}
\usepackage{tabularx}
\usepackage{booktabs}
\usepackage{upquote} 
\graphicspath{{images/}}

\title{3D SEMANTIC SCENE COMPLETION FROM A SINGLE DEPTH IMAGE\\ USING ADVERSARIAL TRAINING}

\name{Yueh-Tung Chen\textsuperscript{1,2}, Martin Garbade\textsuperscript{2}, Juergen Gall\textsuperscript{2}\thanks{The work has been funded by the Deutsche Forschungsgemeinschaft (DFG, German Research Foundation) GA 1927/2-2 (FOR 1505 Mapping on Demand).}}
\address{\textsuperscript{1}b-it, RWTH Aachen University, 52062 Aachen, Germany\\
    \textsuperscript{2}Institute of Computer Science, University of Bonn, 53115 Bonn, Germany\\
    \small{yuehtung.chen@gmail.com, \{garbade, gall\}@iai.uni-bonn.de}
}

\begin{document}
\ninept
\maketitle
\begin{abstract}
We address the task of 3D semantic scene completion, $i.e.$, given a single depth image, we predict the semantic labels and occupancy of voxels in a 3D grid representing the scene. 
In light of the recently introduced generative adversarial networks (GAN), our goal is to explore the potential of this model and the efficiency of various important design choices. Our results show that using conditional GANs outperforms the vanilla GAN setup. We evaluate these architecture designs on several datasets. Based on our experiments, we demonstrate that GANs are able to outperform the performance of a baseline 3D CNN in case of clean annotations, but they suffer from poorly aligned annotations.
\end{abstract}
\section{Introduction}
\label{sec:intro}
Semantic scene completion is a combined task of semantic segmentation and shape completion and discovers the hidden information that is present in a 3D scene. For instance, a depth sensor can only capture information from object surfaces that are visible. Most of the geometric and semantic information of the 3D scene is, however, occluded by the objects themselves. As humans, we can estimate the geometry of objects even in the occluded area from experience, providing us instantly an effective model of the 3D scene surrounding us. 3D semantic scene completion tries to achieve the same goal. Given a single depth image, the goal is to predict the entire 3D geometry of all objects in the scene including the occluded areas. The technique has high potential in many areas ranging from domestic robotics and autonomous vehicles to health-care systems. Without the ability to predict full 3D geometry behind the visible surfaces, robots have to exhaustively explore the occluded space, which is not efficient \cite{scopelliti2005robots}. 

%

3D semantic scene completion has become a popular research problem recently.
Some prior works considered completing and labeling 3D scenes as a combined task, but they used separate modules for feature extraction and context modeling \cite{zheng2013beyond,kim20133d,blaha2016large}. 
Song et al.~\cite{song2017semantic} pioneered in applying deep learning to semantic scene completion. They proposed a 3D convolutional network  that leverages dilated convolutions \cite{chen2018deeplab} as well as skip connections \cite{he2016deep}. Also, this work has been extended by adding a second input stream which contains the 2D semantic labels from RGB images \cite{garbade2018two,liu2018seeandthink}.
In \cite{dai2018scancomplete}, the authors proposed a coarse-to-fine 3D fully convolutional network for processing 3D scenes with arbitrary spatial extents and capturing both local details and the global structure of the scenes. 

At the same time, generative adversarial networks have also received a lot of attention. An adversarial approach to learn a deep generative model was first proposed by Goodfellow et al.~\cite{goodfellow2014generative} and has then been further extended by using conditioning variables in generative adversarial networks \cite{dosovitskiy2015learning}.
Luc et al.~\cite{luc2016semantic} applied GANs to 2D semantic segmentation by adding an adversarial network after a segmentation network to discriminate segmentation maps coming from either the ground truth or the segmentation network. 
Some prior works applied GANs to the 3D space, but they focus on either single object reconstruction \cite{wu2016learning,yang2018dense} or dealing with a scene as a composition of objects \cite{yang2018learning}. 


Wang et al.~\cite{wang2018adversarial} recently proposed a first approach to use generative adversarial networks for 3D semantic scene completion. They use two encoder networks producing a compressed latent state vector of the input depth image and the ground truth volume. Moreover, they propose to use multiple discriminator networks attached to both the output of the encoder networks and the output of a subsequent decoder network. The approach, however, has the disadvantage that the encoder of the depth image differs from the encoder for the voxelized ground truth. As a result, the encoders discard too much information to match the different representations yielding a substantial loss in information.


In this paper, we therefore present an approach that can be combined with any 3D convolutional network that does not suffer from the loss of information. In particular, we propose a conditional generative adversarial network (GAN) to predict semantic labels and occupancy in 3D space simultaneously. We thoroughly evaluate the proposed approach and compare it with a standard generative adversarial network as well as in combination with a local adversarial loss. We observe that the conditional generative adversarial network performs best, but also that generative adversarial networks struggle if the ground truth is not well aligned with the depth data as in NYU Kinect.

\section{Semantic Scene Completion with GANs}
\label{sec:ssc-gan}

Inspired by the successful application of GANs in other domains, we introduce a novel model to perform semantic scene completion using GANs. 

\subsection{Network Architecture}
Our model takes a 3D Truncated Signed Distance Function (TSDF) extracted from a depth image as input and predicts the fully voxelized 3D scene by a generator network. In this work, we use the network architecture SSCNet \cite{song2017semantic} as our generator network. Since the last layer of the generator network is a softmax layer, the output is a probability map over $C$ classes of size $H \times W \times D$, where $H$, $W$, and $D$ are the height, width, and depth of the 3D volume, respectively. The aim of the discriminator network is to distinguish a generated 3D volume from a ground truth volume. To this end, we transform a ground truth sample of the training data to a volume of the same size ($C \times H \times W \times D$) using one-hot encoding. 
%
Although the discriminator network might easily distinguish the ground truth and the generated volume by detecting whether the volume consists of zeros and ones (one-hot encoding) or values between zero and one, Luc et al.~\cite{luc2016semantic} have shown that this encoding mechanism does not strongly affect the performance of the discriminator network.

\begin{figure}[tb]
  \centering
  \centerline{\includegraphics[width=8.5cm]{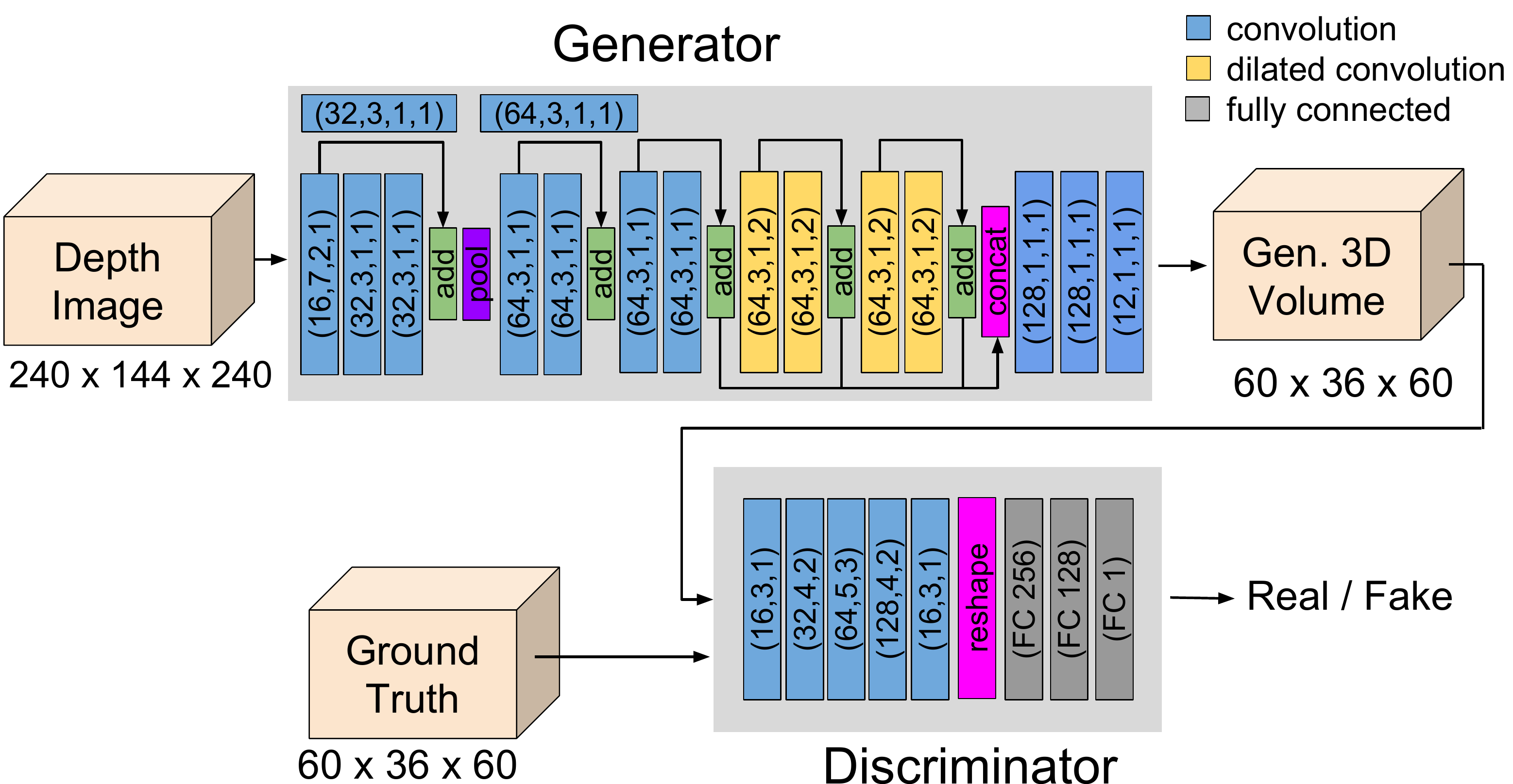}}
\caption{\textbf{Proposed network architecture.} The generator network takes a depth image as input and predicts a 3D volume. The discriminator network takes either the generated 3D volume or the ground truth volume as input and classifies them as real or fake. 
The parameters of each layer are shown as (number of filters, kernel size, stride) in the case of convolutions and as (number of output channels) in the case of fully connected layers.}
\label{fig:GAN}
\end{figure}

Following the design of Wu et al.~\cite{wu2016learning}, the discriminator network consists of several convolutional blocks. Each block comprises a convolution layer with a 3D kernel, 
a normalization layer, and a leaky ReLU activation layer. 
The output of the last convolutional layer with the size of $5\times3\times5\times16$ is reshaped to a vector of 1200 dimensions. After that, it is processed by three fully-connected layers with output sizes of 256, 128 and 1, respectively. Hence, the final logit is a binary indicator to determine whether the predicted volumetric data is generated or sampled from the ground truth data. An overview of our proposed architecture is shown in Figure \ref{fig:GAN}.

\subsection{Loss Function}
We propose to use a hybrid loss function that is a weighted sum of two terms. The first term is a multi-class cross-entropy loss that is used for the generator to predict the right class label at each voxel location independently. We use $g(x)$ to denote the class probability map over the $C$ classes for the volume $H \times W \times D$, which is produced by the generator network.

The second loss term is based on the output of the discriminator network. This loss term is large if the discriminator can differentiate between the predictions of the generator network and the ground truth label maps. We use $d(x,y) \in [0,1]$ to represent the probability with which the discriminator network predicts that $y$ is the ground truth label map of $x$, as opposed to being a label map produced by the generator network $g(\cdot)$.
Given a dataset of $N$ training images $x_n$ and a corresponding 3D ground truth volume $y_n$, we define the loss as:
\begin{equation}\label{eqn:ownGAN}
\begin{aligned}
\mathcal{L}_{GAN}&(\theta_g, \theta_d) = \sum_{n=1}^{N}\mathcal{L}_{mce}(g(x_n),y_n) \\
    &- \lambda[\mathcal{L}_{bce}(d(x_n,y_n),1) + \mathcal{L}_{bce}(d(x_n,g(x_n)),0)]
\end{aligned}
\end{equation}
where $\theta_g$ and $\theta_d$ denote the parameters of the generator and discriminator network, respectively. The multi-class cross-entropy loss for prediction $\hat{y}$ is given by:
\begin{equation}
    \mathcal{L}_{mce}(\hat{y},y) = -\sum_{i=1}^{H \times W \times D}\sum_{c=1}^{C} y_{ic}ln{\hat{y}_{ic}}
\end{equation}
which equals the negative log-likelihood of the target ground truth volume $y$ in a one-hot encoding representation. Similarly, the binary cross-entropy loss is denoted as:
\begin{equation}
    \mathcal{L}_{bce}(\hat{z},z) = -[z ln\hat{z} + (1-z) ln(1-\hat{z})].
    \label{eqn:bce}
\end{equation}
We then minimize the loss according to the parameters $\theta_g$ of the generator network, while maximizing it with respect to the parameters $\theta_d$ of the discriminator network.



\subsection{Conditional GANs}
\label{ssec:cgan}
Conditional GANs have been recently proposed in the literature. Since these works deal with different tasks ($e.g.$ 2D image generation), it is helpful to examine their potential for 3D semantic scene completion.

Using a conditional GAN, the output of the discriminator $d(x, y)$ is conditioned on the input $x$ which we denote as $d(x, y|x)$. This leads to the following new objective function: 
\begin{equation}\label{eqn:owncGAN}
\begin{aligned}
    &\mathcal{L}_{cGAN}(\theta_g, \theta_d) = \sum_{n=1}^{N}\mathcal{L}_{mce}(g(x_n),y_n) \\
    &\quad\; - \lambda[\mathcal{L}_{bce}(d(x_n,y_n|x_n),1) + \mathcal{L}_{bce}(d(x_n,g(x_n)|x_n),0)].
\end{aligned}
\end{equation}
In practice, we achieve the conditioning by concatenating the input depth image $x_n$ with the two kinds of inputs which are fed into the discriminator network respectively.

\subsection{Local Adversarial Loss}
\label{ssec:ll}
A key observation for the discriminator network is that it should learn to model the input sample features equally within the whole input space. When we train a single strong discriminator network, the generator network tends to overemphasize certain parts of the  features to fool the current discriminator network. In other words, any local cube sampled from the input samples should have similar statistics as a real ground truth cube. Therefore, the idea of a local adversarial loss was proposed to overcome this problem for 2D images \cite{shrivastava2017learning}. Here, however, we extend this trick to the 3D domain. Rather than defining a global discriminator network, we can define a discriminator network that classifies each voxel separately. This division strategy not only enhances the capacity of the discriminator network, but also provides more samples per input volume for learning. 

In practice, we design the discriminator network to be a fully convolutional network that outputs the same dimension $C \times H \times W \times D$ as the input. Instead of using fully connected layers to reduce the output into a single probability value, we upsample the output to match the ground truth dimensions. Since the discriminator has shrunk the input volume within the middle three layers by the factor of 12, we again upsample the volume by the factor of 12 using trilinear upsampling. We then calculate the loss term with binary cross-entropy.

\section{Experiments}
\label{sec:exper}
We implement our network architecture in PyTorch \cite{paszke2017automatic} and use a batch size of 4. For our generator network, we use a SGD optimizer with weight decay of 0.0005 and learning rate of 0.01. For the discriminator network, we use an Adam optimizer with a learning rate of 0.0001. Besides, label smoothing is applied for improving the training process in all the experiments \cite{salimans2016improved}. 
We perform some experiments to determine the optimal value for the loss weight paramter $\lambda$ in \eqref{eqn:ownGAN} and \eqref{eqn:owncGAN}. It turns out that $\lambda = 1$ performs best.


We separate our evaluation results mainly in two parts: Semantic scene completion (SSC) and scene completion (SC). While scene completion only considers whether a voxel is occupied or empty, semantic scene completion also evaluates whether an occupied voxel is given the correct semantic label. As in \cite{song2017semantic} we measure the precision, recall, and Jaccard index (IoU) for scene completion and the average (avg.) of the IoU across all categories for semantic scene completion. 

\subsection{Evaluation on NYU Depth v2}
\label{ssec:nyu}
We first evaluate our models on the NYU Depth v2 dataset \cite{silberman2012indoor}, an indoor scene dataset which contains 1449 depth scenes captured by a Kinect device (795 for training, 654 for testing). The annotations consist of 33 CAD models belonging to 7 different categories which have been fit into the scene by human annotators and finally voxelized. Since the CAD models do not perfectly align with the real objects, we also use depth maps generated from the projections of the 3D annotations for training as described in \cite{firman2016structured}. For testing, we consider two test sets: NYU Kinect consisting of depth images captured by the Kinect sensor and NYU CAD consisting of rendered depth images generated by projecting the annotated CAD models. Tables~\ref{table:nyukinect} and \ref{table:nyucad} show the results for four different design choices. Firstly, we examine using the standard GANs vs.\ using conditional GANs which we denote as \textquotesingle GAN\textquotesingle{}  and \textquotesingle cGAN\textquotesingle. Secondly, the usage of the global adversarial loss vs.\ the local adversarial loss is examined, denoted as \textquotesingle GL\textquotesingle{}  and \textquotesingle LL\textquotesingle{}, respectively. 

Table \ref{table:nyukinect} shows the results on the NYU Kinect test set. In case of the global adversarial loss, the accuracy for semantic scene completion increases from 20.6\% to 22.7\% when switching from the GAN to the conditional GAN. Applying the local adversarial loss decreases the accuracy by -1.2\% for the cGAN and by -0.7\% for the GAN. 
For scene completion, the IoU decreases between -0.4\% and \mbox{-1\%} if the conditional GAN is used instead of the GAN. Applying the local adversarial loss decreases the performance between -0.1\% and -0.7\%.
Overall our model performs worse than the baseline \cite{song2017semantic} for semantic scene completion, which achieves 24.7\% on NYU Kinect, whereas our best model only achieves 22.7\%. This suggests that our model is less robust to noise in the test data of NYU Kinect than \cite{song2017semantic}.
Due to the inaccurate annotations of the data in NYU Kinect, we conclude that NYU Kinect is not suited to examine the potential of adversarial learning. We therefore focus on the results on the NYU CAD test set.

Table \ref{table:nyucad} shows the results on the NYU CAD test set. The accuracy for semantic scene completion improves by around +2\% (from 39.8\% to 42.0\%) by the usage of the conditional GAN and further increases slightly to 42.3\% by applying the local adversarial loss. The approach outperforms the baseline \cite{song2017semantic} not only for semantic scene completion, but also for scene completion. In this case, the best accuracy is also achieved by the conditional GAN with local adversarial loss (74.9\%).

In summary, the conditional GAN outperforms the standard GAN setup for both NYU test settings.
Local and global adversarial loss, however, perform differently on different test settings. While the local adversarial loss slightly improves the accuracy on NYU CAD, it decreases the accuracy on NYU Kinect. Since the local adversarial loss tends to put more emphasis on the finer local detail of each voxel, it is more sensitive to the inaccurate annotations of NYU Kinect. 

We also observe that \cite{song2017semantic} achieves a very high recall but a low precision for scene completion. In other words, this method tends to predict more occupied voxels than our approaches, which results in cluttered scene completions. 
Nevertheless, on NYU CAD, our models outperform the baseline \cite{song2017semantic} by a large margin for both scene completion and semantic scene completion. As shown in Table \ref{table:nyucad}, our model SSC-cGAN-LL achieves 42.3\% accuracy for semantic scene completion and outperforms the approach \cite{song2017semantic} by +4.7\%. For scene completion, our model outperforms \cite{song2017semantic} by +4.6\% IoU.

\begin{figure}[tb]
  \centering
  \centerline{\includegraphics[width=8.5cm]{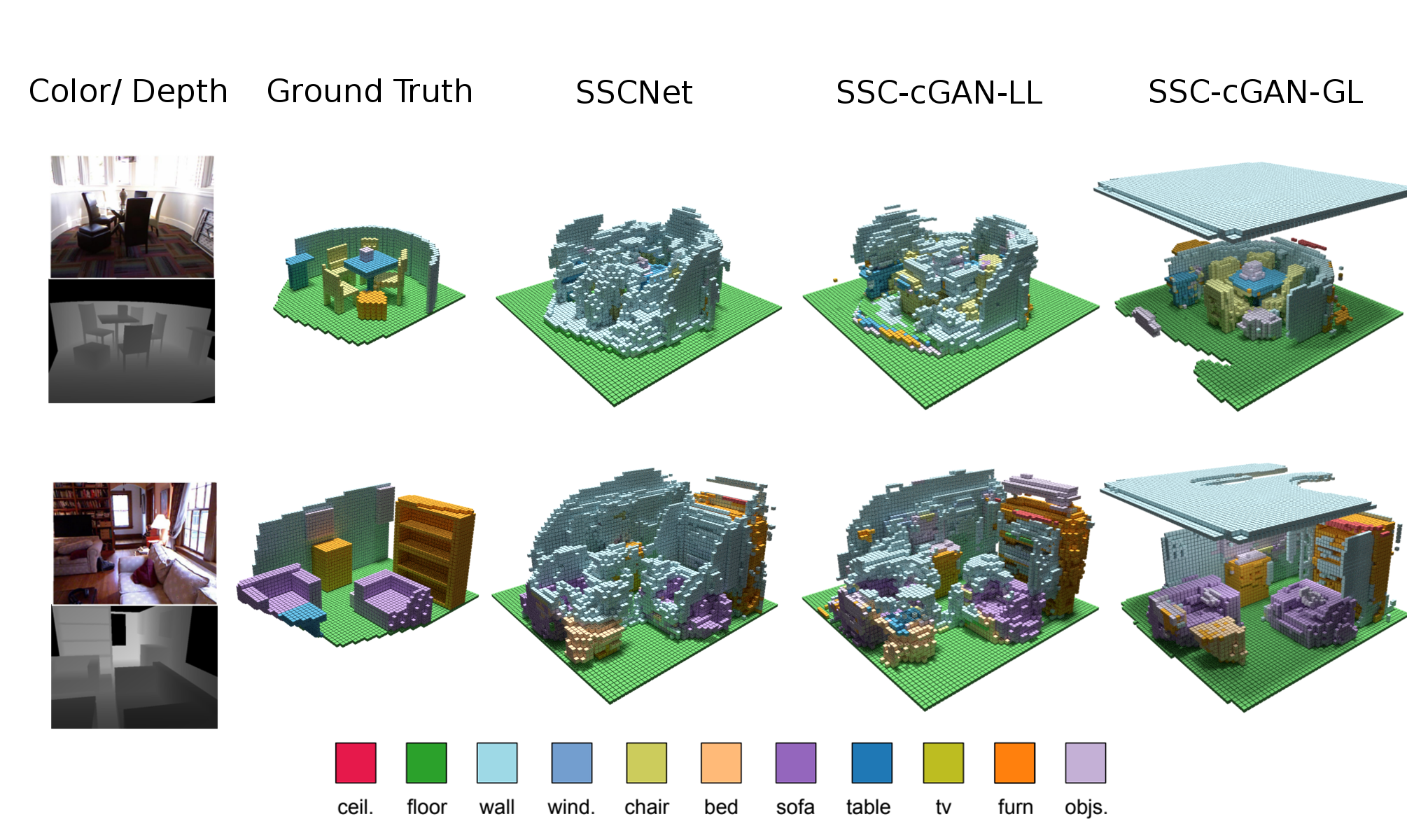}}
\caption{\textbf{Qualitative results on NYU CAD.} The first three columns show the input depth image with its corresponding color image, ground truth volume, and the results obtained by \cite{song2017semantic}. The fourth and fifth columns show the results obtained by our approaches.}
\label{fig:visualResult}
\end{figure}

Apart from the quantitative results, we provide also some qualitative results in Figure \ref{fig:visualResult} to visualize the effect of applying different GAN models. On the one hand, we can observe that the model using the global GAN loss (SSC-cGAN-GL) suffers from partial mode collapse, meaning it constantly generates labeled voxels in parts of the scene, in particular for ceilings and walls. Using a conditional GAN together with the local adversarial loss seems to significantly reduce this problem. On the other hand, the predicted scenes from SSC-cGAN-GL are visually more plausible since they display more fine structure of the objects, tend to contain more empty voxels and therefore look less cluttered. Thus, we propose SSC-cGAN-GL as the most effective model, due to the fact that it performs comparably to the model with the highest accuracy (SSC-cGAN-LL), while being able to generate more realistic results.


Finally, we compare our approach to the state-of-the-art. On NYU Kinect, the proposed approaches perform worse than the baseline \cite{song2017semantic} for semantic scene completion, but better for scene completion. The only approach that fairly outperforms the baseline is \cite{zhang2018efficient}. All the other approaches use either an additional modality (RGB images) \cite{garbade2018two}, pretrain on SUNCG \cite{song2017semantic}, or do both \cite{liu2018seeandthink}. On NYU CAD, we outperform \cite{song2017semantic} even if they pretrain on SUNCG and perform competitively with \cite{garbade2018two}.

\begin{table}[tb]
\centering
\begin{tabular}{l|ccc|c}
\toprule
NYU Kinect & \multicolumn{3}{c|}{SC} & SSC \\ 
\hline
method & prec. & recall & IoU & avg. \\ \hline
SSCNet \cite{song2017semantic} & 57.0 & 94.5 & 55.1 & 24.7 \\
Zhang et al. \cite{zhang2018efficient} & 71.9 & 71.9 & 56.2 & 26.7\\
Garbade et al. \cite{garbade2018two} & 65.6 & 87.2 & 60.0 & 34.1\\
Liu et al.\textsuperscript{*} \cite{liu2018seeandthink} & 67.3& 85.8 & \textbf{60.6} & \textbf{34.4}\\
SSCNet\textsuperscript{*} \cite{song2017semantic} & 59.3 & 92.9 & 56.6 & 30.5 \\
\hline
SSC-GAN-GL & 65.3 & 84.8 & 58.2 & 20.6 \\
SSC-GAN-LL & 64.5 & 85.9 & 58.1 & 19.9 \\
SSC-cGAN-GL & 63.1 & 87.8 & 57.8 & 22.7 \\
SSC-cGAN-LL & 64.0 & 84.8 & 57.1 & 21.5 \\

\bottomrule                
\end{tabular}
\caption{Comparison of models for semantic scene completion on the NYU Kinect dataset. \textsuperscript{*} denotes that the network is trained on SUNCG and fine-tuned on NYU.}
\label{table:nyukinect}
\end{table}

\begin{table}[tb]
\centering
\begin{tabular}{l|ccc|c}
\toprule
NYU CAD & \multicolumn{3}{c|}{SC} & SSC \\ 
\hline
method & prec. & recall & IoU & avg. \\ \hline
SSCNet \cite{song2017semantic} & 75.0 & 92.3 & 70.3 & 37.6\\
SSCNet\textsuperscript{*} \cite{song2017semantic} & 75.4 & 96.3 & 73.2 & 40.0\\
Garbade et al. \cite{garbade2018two} & 81.6 & 92.4 & \textbf{76.1} & \textbf{46.2}\\
\hline
SSC-GAN-GL & 81.1 & 90.6 & 74.8 & 39.8 \\
SSC-GAN-LL & 80.6 & 91.3 & 73.9 & 40.6 \\
SSC-cGAN-GL & 80.7 & 91.1 & 74.8 & 42.0 \\
SSC-cGAN-LL & 81.0 & 91.0 & 74.9 & 42.3 \\
\bottomrule                
\end{tabular}
\caption{Comparison of models for semantic scene completion on the NYU CAD dataset~\cite{firman2016structured}. \textsuperscript{*} denotes that the network is trained on SUNCG and fine-tuned on NYU.}
\label{table:nyucad}
\end{table}

\subsection{Evaluation on SUNCG}
\label{ssec:suncg}
The SUNCG dataset \cite{song2017semantic} is a synthetic dataset, which provides a large amount of training data with rendered depth images and volumetric ground truth. It contains 45,622 different scenes with realistic room and furniture layouts. However, due to the high computational cost (around 15 days for 10 epochs) of the training procedure, we only run our SSC-cGAN-GL model on SUNCG. The result is shown in Table \ref{table:suncg}. Our model SSC-cGAN-GL outperforms the baseline \cite{song2017semantic} by a large margin of +10\% for semantic scene completion. For scene completion the IoU increases from 72.9\% to 78.1\% with both higher precision and recall. As a result, we conclude that the GAN structure benefits from the large amount of training samples provided by SUNCG.

Compared to other recent approaches, our approach achieves a lower accuracy. However, since our approach is orthogonal to the approaches of \cite{zhang2018efficient,liu2018seeandthink}, we expect that they could be combined and potentially benefit each other. \cite{wang2018adversarial} also used an adversarial learning approach in combination with an encoder-decoder network. However, they do not follow the original evaluation protocol. Instead of reporting numbers on the SUNCG test set, they perform a 10-fold cross validation using random splits. For comparison to the state-of-the-art, we follow the standard evaluation protocol \cite{song2017semantic}.

\begin{table}[htb]
\centering
\begin{tabular}{l|ccc|c}
\toprule
SUNCG & \multicolumn{3}{c|}{SC} & SSC \\ 
\hline
method & prec. & recall & IoU  & avg. \\ \hline
SSCNet \cite{song2017semantic} & 79.8 & 89.5 & 72.9  & 45.0 \\
Wang et al. \cite{wang2018adversarial} \textsuperscript{*} & - & - & - & 51.4 \\ 
Zhang et al. \cite{zhang2018efficient} & 92.6 & 90.4 & \textbf{84.5} & \textbf{70.5} \\ 
Liu et al. \cite{liu2018seeandthink} & 80.7 & 96.5 & 78.5 & 64.3 \\ \hline
SSC-cGAN-GL & 83.4 & 92.4 & 78.1 & 55.6 \\
\bottomrule                
\end{tabular}
\caption{Comparison with the state-of-the-art on the SUNCG dataset. \textsuperscript{*} denotes that the model uses different training / testing splits and performs a 10-fold cross-validation \cite{wang2018adversarial}}
\label{table:suncg}
\end{table}

\subsection{Loss Behaviour of the Discriminator}
\label{ssec:lossbehave}
\begin{figure}[tb]
  \centering
  \centerline{\includegraphics[width=8.5cm]{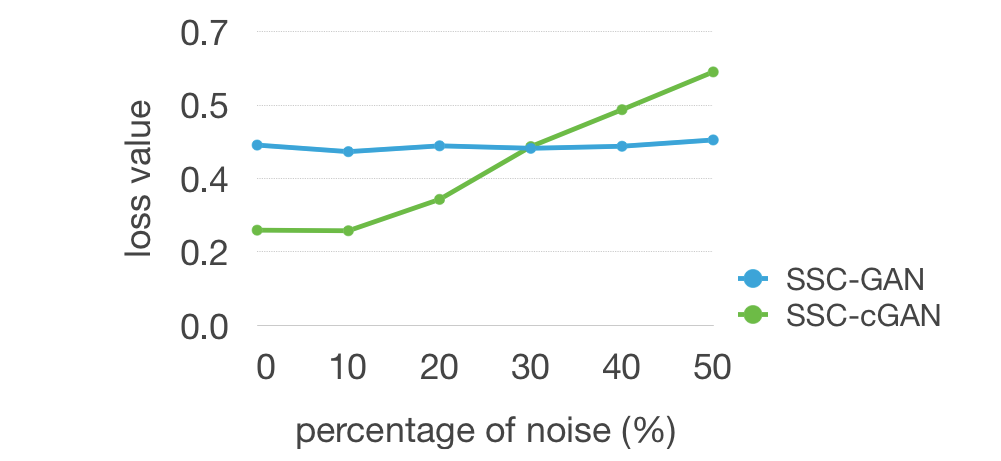}}
\caption{Comparison of the loss behaviour of the discriminator networks.}
\label{fig:disCurve}
\end{figure}

We design an experiment that allows us to assess whether the discriminators show the expected behaviour of producing high losses for unrealistic scene inputs. Therefore, we gradually add noise to the ground truth samples of the NYU CAD test data and feed it as input to the trained discriminator networks. We simulate the noise by randomly changing voxel labels in the occluded space. While increasing the percentage of noise, we calculate the binary cross-entropy loss value using~\eqref{eqn:bce}. 
As one can see from Figure~\ref{fig:disCurve}, the loss curve for SSC-GAN remains stable whereas it increases for SSC-cGAN. This suggests that the conditional GAN model behaves in the expected way while the standard GAN is insensitive to the noise. 


\section{Conclusion}
\label{sec:conclusion}

We presented a novel GAN architecture to perform 3D semantic scene completion and evaluated two variations, namely conditional GANs and a local adversarial loss. 
The results show that the conditional GAN improves the network performance on both test sets, while the local adversarial loss only improves the performance on NYU CAD but not on NYU Kinect. In comparison to the baseline \cite{song2017semantic}, our models yield a significant improvement on NYU CAD.
On SUNCG our models outperform the baseline by a large margin. If we compare the results qualitatively, the proposed model SSC-cGAN-GL produces significantly more realistic appearing scenes than the baseline or the local adversarial loss.


\bibliographystyle{IEEEbib}
\bibliography{main}

\end{document}